\RequirePackage{fix-cm}
\documentclass[natbib]{svjour3}
\smartqed  % flush right qed marks, e.g. at end of proof
\usepackage{graphicx}
\usepackage{latexsym,algorithm,algorithmic}
\usepackage{times}
\usepackage{amsmath,amssymb,mathtools}
\usepackage{mathptmx}      % use Times fonts if available on your TeX system
\begin{document}

\title{Efficient Inference and Learning in a Large Knowledge Base}
%Insert your title here%\thanks{Grants or other notes

\subtitle{Reasoning with Extracted Information using a Locally Groundable First-Order Probabilistic Logic}


\author{William Yang Wang
           \and Kathryn Mazaitis
           \and Ni Lao
           \and Tom Mitchell
           \and William W. Cohen
}


\institute{W. Y. Wang, K. Mazaitis, T. Mitchell, W. W. Cohen  \at
              School of Computer Science, Carnegie Mellon University\\
             5000 Forbes Ave.\\
    Pittsburgh, PA 15213, U.S.A.\\
              Tel.: +1-412-268-7664\\
              Fax: +1-412-268-2205\\
              \email{\{ww, krivard, tom.mitchell, wcohen\}@cs.cmu.edu}           %  \\
           \and
           N. Lao \at
           Google Inc.\\
          1600 Amphitheatre Parkway\\
          Mountain View, CA 94043, U.S.A.\\
         \email{nlao@google.com}
}

\newcommand{\wc}[1]{{\bf{[wc: #1]}}} 
\newcommand{\trm}[1]{\emph{#1}}
\newcommand{\ct}[1]{{\it ct: {#1}}}
\newcommand{\uncite}[1]{}
\newcommand{\weightvec}{\textbf{w}}
\newcommand{\edge}[2]{{u\rightarrow{}v}}
\newcommand{\edgeuv}{{\edge{u}{v}}}

\date{Received: January 10, 2014 / Accepted: date}

\maketitle

\begin{abstract}
One important challenge for probabilistic logics is reasoning with
very large knowledge bases (KBs) of imperfect information, such as
those produced by modern web-scale information extraction systems.
One scalability problem shared by many probabilistic logics is that
answering queries involves ``grounding'' the query---i.e., mapping it
to a propositional representation---and the size of a ``grounding''
grows with database size.  To address this bottleneck, we present a
first-order probabilistic language called ProPPR in which that
approximate ``local groundings'' can be constructed in time
\emph{independent} of database size.  Technically, ProPPR is an
extension to \trm{stochastic logic programs} (SLPs) that is biased
towards short derivations; it is also closely related to an earlier
relational learning algorithm called the \trm{path ranking algorithm}
(PRA).  We show that the problem of constructing proofs for this logic
is related to computation of \trm{personalized PageRank} (PPR) on a
linearized version of the proof space, and using on this connection,
we develop a proveably-correct approximate grounding scheme, based on
the PageRank-Nibble algorithm.  Building on this, we develop a fast
and easily-parallelized weight-learning algorithm for ProPPR.  In
experiments, we show that learning for ProPPR is orders magnitude
faster than learning for Markov logic networks; that allowing mutual
recursion (joint learning) in KB inference leads to
improvements in performance; and that ProPPR can learn weights for a
mutually recursive program with hundreds of clauses, which define
scores of interrelated predicates, over a KB containing one million
entities.

\end{abstract}

%\begin{figure}
%  \includegraphics{example.eps}
%\end{figure}
%\begin{figure*}
%  \includegraphics[width=0.75\textwidth]{example.eps}
%\end{figure*}
%\begin{table}
%\begin{tabular}{lll}
%\hline\noalign{\smallskip}
%\end{tabular}
%\end{table}

\section{Introduction} 
 
While probabilistic logics are useful for many important tasks
\citep{lowd2007efficient,fuhr1995probabilistic,poon2007joint,poon2008joint};
in particular, such logics would seem to be well-suited for inference
with the ``noisy'' facts that are extracted by automated systems from
unstructured web data.  While some positive results have been obtained
for this problem \citep{CohenTOIS2000}, most probabilistic first-order
logics are not efficient enough to be used for inference on the very
large broad-coverage KBs that modern information extraction systems
produce \citep{suchanek2007yago,DBLP:conf/aaai/CarlsonBKSHM10}.  One key problem is that queries are typically
answered by ``grounding'' the query---i.e., mapping it to a
propositional representation, and then performing propositional
inference---and for many logics, the size of the ``grounding'' can be
extremely large for large databases.  For instance, in probabilistic
Datalog \citep{fuhr1995probabilistic}, a query is converted to a
structure called an ``event expression'', which summarizes all
possible proofs for the query against a database; in ProbLog~\citep{de2007problog} 
and MarkoViews~\citep{jha2012probabilistic}
similar structures are created, encoded more compactly with binary
decision diagrams (BDDs); in probabilistic similarity logic (PSL) an
intentional probabilistic program, together with a database, is
converted to constraints for a convex optimization problem; and in
Markov Logic Networks (MLNs) \citep{RichardsonMLJ2006}, queries are
converted to a (propositional) Markov network.  In all of these cases,
the result of this ``grounding'' process can be large.

\begin{figure}
%\wc{make the A,B in this figure lower-case for the ILP people!}
%WW: done.
\centerline{\includegraphics[scale=0.35]{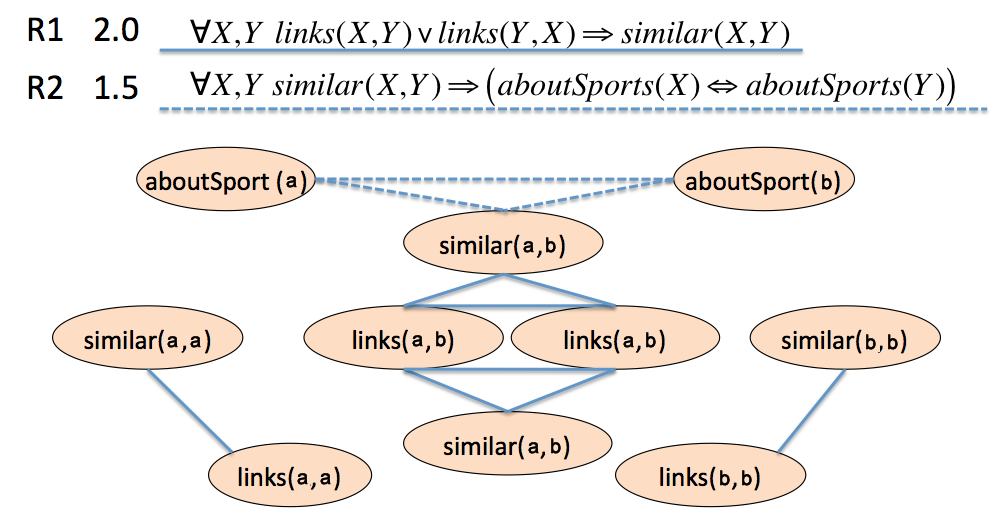}}
\caption{A Markov logic network program and its grounding relative to
  two constants $a,b$.  (Dotted lines are clique potentials associated
  with rule R2, solid lines with rule R1.)}
\label{fig:mln}
\end{figure}

As concrete illustration of the ``grounding'' process,
Figure~\ref{fig:mln} shows a very simple MLN and its grounding over a
universe of two web pages $a$ and $b$. (Here the grounding is
query-independent.).  In MLNs, the result of the grounding is a Markov
network which contains one node for every atom in the Herbrand base of
the program---i.e., the number of nodes is $O(n^k)$ where $k$ is the
maximal arity of a predicate and $n$ the number of database constants.
However, even a grounding of size that is only linear in the number of
facts in the database, $|DB|$, would impractically large for inference
on real-world problems.  Superficially, it would seem that groundings
must inherently be $o(|DB|)$ for some programs: in the example, for
instance, the probability of \textit{aboutSport(x)} must depend to
some extent on the entire hyperlink graph (if it is fully connected).
However, it also seems intuitive that if we are interested in
inferring information about a specific page---say, the probability of
\textit{aboutSport(d1)}---then the parts of the network only distantly
connected to \textit{d1} are likely to have a small influence.  This
suggests that an \emph{approximate} grounding strategy might be
feasible, in which a query such as \textit{aboutSport(d1)} would be
grounded by constructing a small subgraph of the full network,
followed by inference on this small ``locally grounded'' subgraph.
Likewise, consider learning (e.g., from a set of queries $Q$ with
their desired truth values). Learning might proceed by
locally-grounding every query goal, allowing learning to also take
less than $O(|DB|)$ time.

In this paper, we present a first-order probabilistic language which
is well-suited to such approximate ``local grounding''.  We describe
an extension to \trm{stochastic logic programs} (SLP)
\citep{DBLP:journals/ml/Cussens01} that is biased towards short
derivations, and show that this is related to \trm{personalized
  PageRank} (PPR) \citep{pagerank,SoumenWWW2007} on a linearized
version of the proof space. Based on the connection to PPR, we develop
a proveably-correct approximate inference scheme, and an associated
proveably-correct approximate grounding scheme: specifically, we show
that it is possible to prove a query, or to build a graph which
contains the information necessary for weight-learning, in time
$O(\frac{1}{\alpha\epsilon})$, where $\alpha$ is a reset parameter
associated with the bias towards short derivations, and $\epsilon$ is
the worst-case approximation error across all intermediate stages of
the proof.  This means that both inference and learning can be
approximated in time \emph{independent of the size of the underlying
  database}---a surprising and important result, which leads to a very
scalable inference algorithm.

The ability to locally ground queries has another important
consequence: it is possible to \emph{decompose} the problem of
weight-learning to a number of moderate-size subtasks (in fact, tasks
of size $O(\frac{1}{\alpha\epsilon})$ or less) which are weakly
coupled.  Based on this we outline a parallelization scheme, which in
our current implementation provides an order-of-magnitude speedup in
learning time on a multi-processor machine.

Below, we will first introduce our formalism, and then describe our
weight-learning algorithm.  We next present experimental results on
some small benchmark inference tasks.  We then present experimental
results on a larger, more realistic task: learning to perform accurate
inference in a large KB of facts extracted from the web
\citep{DBLP:conf/emnlp/LaoMC11}.  We finally discuss related work and
conclude.

\section{\underline{Pro}gramming with \underline{P}ersonalized
  \underline{P}age\underline{R}ank (PROPPR)}

\subsection{Inference as Graph Search}

We will now describe our ``locally groundable'' first-order
probabilistic language, which we call ProPPR. 
Inference for ProPPR is based on
a personalized PageRank process over the proof constructed by Prolog's
Selective Linear Definite (SLD) resolution theorem-prover.  To define the
semantics we will use notation from logic programming \citep{Lloyd}.
Let $LP$ be a program which contains a set of definite clauses
$c_1,\ldots,c_n$, and consider a conjunctive query $Q$ over the
predicates appearing in $LP$.  A traditional Prolog interpreter can be
viewed as having the following actions.  First, construct a ``root
vertex'' $v_0$, which is a pair $(Q,Q)$ and add it to an
otherwise-empty graph $G'_{Q,LP}$. (For brevity, we drop the
subscripts of $G'$ where possible.)  Then recursively add to $G'$ new
vertices and edges as follows: if $u$ is a vertex of the form $(Q,
(R_1,\ldots,R_k))$, and $c$ is a clause in $LP$ of the form \( R'
\leftarrow S'_1,\ldots,S'_\ell \), and $R_1$ and $R'$ have a most
general unifier $\theta=mgu(R_1,R')$, then add to $G'$ a new edge
$\edgeuv$ where \( v = (Q\theta,
(S'_1,\ldots,S'_\ell,R_2,\ldots,R_k)\theta) \).  
Let us call $Q\theta$
the \trm{transformed query} and
$(S'_1,\ldots,S'_\ell,R_2,\ldots,R_k)\theta$ the \trm{associated
  subgoal list}.  If a subgoal list is empty, we will denote it by
$\Box$. Here $Q\theta$ denotes the result of applying the substitution
$\theta$ to $Q$; for instance, if $Q = about(a,Z)$
and $\theta = \{Z = fashion \}$, then $Q\theta$
is $about(a,fashion)$.

The graph $G'$ is often large or infinite so it is not constructed
explicitly. Instead Prolog performs a depth-first search on $G'$ to
find the first \trm{solution vertex} $v$---i.e., a vertex with an
empty subgoal list---and if one is found, returns the transformed
query from $v$ as an answer to $Q$.  

\begin{figure*}
\centerline{\includegraphics[scale=0.3]{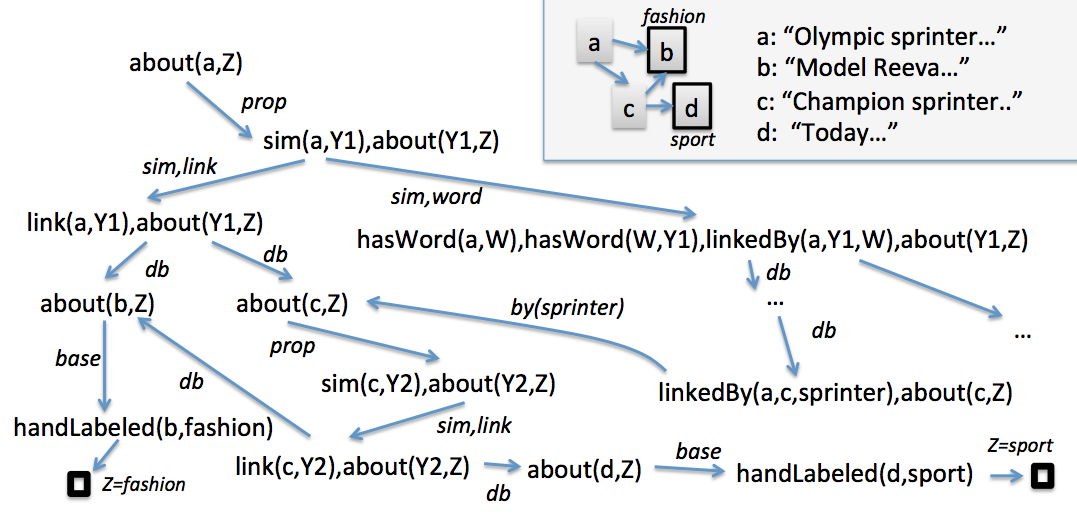}}
\caption{A partial proof graph for the query \textit{about(a,Z)}.  The
  upper right shows the link structure between documents $a,b,c$, and
  $d$, and some of the words in the documents.  Restart links are not
  shown.}
\label{fig:proof}
\end{figure*}

Table~\ref{tab:proppr} and Figure~\ref{fig:proof} show a simple Prolog
program and a proof graph for it. The annotations after the hashmarks
and the edge labels in the proof graph will be described below in more
detail: briefly, however, we will associated with each use of a clause
$c$ a \trm{feature vector} $\phi$, which is computed from the binding
to the variables in the head of $c$.  For instance, applying the
clause \textit{``sim(X,Y):-links(X,Y)''} always yields a vector $\phi$
that has unit weight on (the dimensions corresponding to) the two
ground atoms \textit{sim} and \textit{link}, and zero weight
elsewhere; likewise, applying the clause
\textit{``linkedBy(X,Y),W:-''} to the goal
\textit{linkedBy(a,c,sprinter)} yields a vector $\phi$ that has unit
  weight on the atom \textit{by(sprinter)}.

For conciseness, in Figure~\ref{fig:proof} only the subgoals
$R_1,\ldots,R_k$ are shown in each node $u=(Q,(R_1,\ldots,R_k))$.
Given the query $Q=\textit{about(a,Z)}$, Prolog's depth-first search
would return $Q=\textit{about(a,fashion)}$.  Note that in this proof
formulation, the nodes are \emph{conjunctions} of literals, and the
structure is, in general, a digraph (rather than a tree). Also note
that the proof is encoded as a graph, not a hypergraph, even if the
predicates in the LP are not binary: the edges represent a step in the
proof that reduces one conjunction to another, not a binary relation
between entities.

\begin{table}
\caption{A simple program in ProPPR.  See text for explanation.} \label{tab:proppr}
\begin{center}
\begin{tabular}{|ll|}
\hline
about(X,Z) :- handLabeled(X,Z)    &\# base. \\
about(X,Z) :- sim(X,Y),about(Y,Z)   &\# prop. \\
sim(X,Y) :- links(X,Y)                      &\# sim,link. \\
sim(X,Y) :-                                 & \\
~~~~hasWord(X,W),hasWord(Y,W),              & \\
~~~~linkedBy(X,Y,W)                         & \# sim,word.\\
linkedBy(X,Y,W) :- true                     & \# by(W). \\
\hline
\end{tabular}
\end{center}
\end{table}

As an further illustration of the sorts of ProPPR programs that are
possible, some small sample programs are shown in
Figure~\ref{tab:sample}.  Clauses $c_1$ and $c_2$ are, together, a
bag-of-words classifier: each proof of \emph{predictedClass(D,Y)} adds
some evidence for $D$ having class $Y$, with the weight of this
evidence depending on the weight given to $c_2$'s use in establishing
\emph{related(w,y)}, where $w$ and $y$ are a specific word in $D$ and
$y$ is a possible class label. In turn, $c_2$'s weight depends on the
weight assigned to the $r(w,y)$ feature by $\weightvec$, relative to
the weight of the restart link.\footnote{The existence of the restart
  link thus has another important role in this program, as it avoids a
  sort of ``label bias problem'' \uncite{LaffertyML2001} in which
  local decisions are difficult to adjust.}  Adding $c_3$ and $c_4$ to
this program implements label propagation, and adding $c_5$ and $c_6$
implements a sequential classifier.  These examples show that ProPPR
allows many useful heuristics to be encoded as programs.

\begin{table*}
\caption{Some more sample ProPPR programs. $LP=\{c_1,c_2\}$ is a bag-of-words classifier (see text).
  $LP=\{c_1,c_2,c_3,c_4\}$ is a recursive label-propagation scheme, in
  which predicted labels for one document are assigned to similar
  documents, with similarity being an (untrained) cosine distance-like
  measure. $LP=\{c_1,c_2,c_5,c_6\}$ is a sequential classifier for
  document sequences.}

\label{tab:sample}

\smallskip

%\begin{tabular}[ll]
\begin{center}
\begin{minipage}[t]{0.4\textwidth}
\begin{tabbing}
$c_1$: \=pred\=ictedClass(Doc,Y) :- \\
\> \>possibleClass(Y),\\
\> \>hasWord(Doc,W),\\
\> \>related(W,Y) \# c1.\\
$c_2$: related(W,Y) :- true,\\
\>\# relatedFeature(W,Y)\\
~\\
~\\
\emph{Database predicates}:\\
\emph{hasWord(D,W): doc $D$ contains word $W$}\\
\emph{inDoc(W,D): doc $D$ contains word $W$}\\
\emph{previous(D1,D2): doc $D2$ precedes $D1$}\\
\emph{possibleClass(Y): $Y$ is a class label}\\
\end{tabbing}
\end{minipage}~~~~~~\begin{minipage}[t]{0.4\textwidth}
\begin{tabbing}
$c_3$: \=pre\=dictedClass(Doc,Y) :- \\
\> \>similar(Doc,OtherDoc),\\
\> \>predictedClass(OtherDoc,Y) \# c3.\\
$c_4:$ similar(Doc1,Doc2) :- \\
\> \>hasWord(Doc1,W),\\
\> \>inDoc(W,Doc2) \# c4.\\
~\\
$c_5:$ predictedClass(Doc,Y) :- \\
\> \>previous(Doc,OtherDoc),\\
\> \>predictedClass(OtherDoc,OtherY),\\
\> \>transition(OtherY,Y) \# c5.\\
$c_6$: transition(Y1,Y2) :- true, \\
\> \# transitionFeature(Y1,Y2) \\
\end{tabbing}
\end{minipage}
\end{center}
%\end{tabular}
\end{table*}

\subsection{From SLPs to ProPPR}

In \trm{stochastic logic programs} (SLPs)
\citep{DBLP:journals/ml/Cussens01}, one defines a randomized procedure
for traversing the graph $G'$, which thus defines a probability
distribution over vertices $v$, and hence (by selecting only solution
vertices) a distribution over transformed queries (i.e. answers)
$Q\theta$.  The randomized procedure thus produces a distribution over
possible answers, which can be tuned by learning to upweight desired
(correct) answers and downweight others.

In past work, the randomized traversal of $G'$ was defined by a
probabilistic choice, at each node, of which clause to apply, based on
a weight for each clause.  We propose two extensions.  First, we will
introduce a new way of computing clause weights, which allows for a
potentially richer parameterization of the traversal process.  We will
associate with each edge $\edgeuv$ in the graph a \trm{feature vector}
$\phi_{\edgeuv}$.  This edge is produced indirectly, by associating
with every clause $c\in{}LP$ a function $\Phi_c(\theta)$, which
produces the vector $\phi$ associated with an application of $c$ using
mgu $\theta$.  As an example, if the last clause of the program in
Table~\ref{tab:proppr} was applied to $(Q,
linkedBy(a,c,sprinter),about(c,Z))$ with mgu
$\theta=\{X=a,Y=c,W=sprinter\}$ then $\Phi_c(\theta)$ would be
$\{by(sprinter)\}$, if we use a set to denote a sparse vector with 0/1
weights.

This feature vector is computed during theorem-proving, and used to
annotate the edge $\edgeuv$ in $G'$ created by applying $c$ with mgu
$\theta$. Finally, an edge $\edgeuv$ will be traversed with
probability \( \Pr(v|u) \propto f(\weightvec,\phi_\edgeuv) \) where
$\weightvec$ is a parameter vector and where $f(\weightvec,\phi)$ is a
weighting function. (Here we use $f(\weightvec,\phi) = \exp(
\weightvec_i\cdot \phi)$, but any differentiable function would be
possible.)  This weighting function now determines the probability of
a transition, in theorem-proving, from $u$ to $v$: specifically,
$\Pr_\weightvec(v|u) \propto f(\weightvec,\phi_\edgeuv)$.  Weights in
$\weightvec$ default to 1.0, and learning consists of tuning these
weights.

The second and more fundamental extension is to add edges in $G'$ from
every solution vertex to itself, and also add an edge from every
vertex to the start vertex $v_0$. We will call this augmented graph
$G_{Q,LP}$ below (or just $G$ if the subscripts are clear from
context).  These links make SLP's graph traversal a \trm{personalized
  PageRank} (PPR) procedure\uncite{pagerank,SoumenWWW2007}, sometimes
known as \trm{random-walk-with-restart}
\citep{DBLP:conf/icdm/TongFP06}.  These links are annotated by another
feature vector function $\Phi_{\mathrm{restart}}(R)$, which is applied to the
leftmost literal $R$ of the subgoal list for $u$ to annotate the edge
$\edge{u}{v}_{0}$.

These links back to the start vertex bias the traversal of the
proof graph to upweight the results of \emph{short proofs}. To see
this, note that if the restart probability $P(v_0|u)=\alpha$ for every
node $u$, then the probability of reaching any node at depth $d$ is
bounded by $(1-\alpha)^d$.

To summarize, if $u$ is a node of the search graph,
$u=(Q\theta,(R_1,\ldots,R_k))$, then the transitions from $u$, and
their respective probabilities, are defined as follows, where $Z$ is
an appropriate normalizing constant:
\begin{itemize}
\item If \( v = (Q\theta\sigma,
  (S'_1,\ldots,S'_\ell,R2,\ldots,R_k)\theta\sigma) \) is a state
  derived by applying the clause $c$ (with mgu $\sigma$), then
\[ \Pr_\weightvec(v|u)=\frac{1}{Z} f(\weightvec,\Phi_c(\theta\circ\sigma))
\]
\item If \( v = v_0 = (Q,Q) \) is the initial state in $G$, then
\[ \Pr_\weightvec(v|u)=\frac{1}{Z} f(\weightvec,\Phi_{\mathrm{restart}}(R_1\theta))
\]
\item If $v$ is any other node, then $\Pr(v|u)=0$.
\end{itemize}

Finally we must specify the functions $\Phi_c$ and $\Phi_{\mathrm{restart}}$.
For clauses in $LP$, the feature-vector producing function
$\Phi_c(\theta)$ for a clause is specified by annotating $c$ as
follows: every clause $c=(R\leftarrow S_1,\ldots,S_k)$ can be
annotated with an additional conjunction of ``feature literals''
$F_1,\ldots,F_\ell$, which are written at the end of the clause after
the special marker ``\#''.  The function $\Phi_c(\theta)$ then returns
a vector $\phi=\{F_1\theta,\ldots,F_\ell\theta\}$, where every
$F_i\theta$ must be ground.

The requirement\footnote{The requirement that the feature literals
  returned by $\Phi_c(\theta)$ must be ground in $\theta$ is not
  strictly necessary for correctness.  However, in developing ProPPR
  programs we noted than non-ground features were usually not what the
  programmer intended.} that edge features $F_i\theta$ are ground is
the reason for introducing the apparently unnecessary predicate
\textit{linkedBy(X,Y,W)} into the program of Table~\ref{tab:proppr}:
adding the feature literal \textit{by(W)} to the second clause for
\textit{sim} would result in a non-ground feature \textit{by(W)},
since $W$ is a free variable when $\Phi_c$ is called.  Notice also
that the weight on the \textit{by(W)} features are meaningful, even
though there is only one clause in the definition of
\textit{linkedBy}, as the weight for applying this clause competes
with the weight assigned to the restart edges.

It would be cumbersome to annotate every database fact, and difficult
to learn weights for so many features. Thus, if $c$ is the unit clause
that corresponds to a database fact, then $\Phi_c(\theta)$ returns a
default value $\phi=\{\textit{db}\}$, where \textit{db} is a special
feature indicating that a database predicate was used.\footnote{If a
  non-database clause $c$ has no annotation, then the default vector
  is $\phi=\{\textit{id(c)}\}$, where \textit{c} is an identifier for
  the clause $c$.}

The function $\Phi_{\mathrm{restart}}(R)$ depends on the functor and arity of
$R$.  If $R$ is defined by clauses in $LP$, then $\Phi_{\mathrm{restart}}(R)$
returns a unit vector $\phi=\{\textit{defRestart}\}$.  If $R$ is a
database predicate (e.g., \textit{hasWord(doc1,W)}) then we follow a
slightly different procedure, which is designed to ensure that the
restart link has a reasonably large weight even with unit feature
weights: we compute $n$, the number of possible bindings for $R$, and
set $\phi[\textit{defRestart}] = n\cdot{}\frac{\alpha}{1-\alpha}$,
where $\alpha$ is a global parameter.  This means that with unit
weights, after normalization, the probability of following the restart
link will be $\alpha$.

Putting this all together with the standard iterative approach to
computing personalized PageRank over a graph \citep{pagerank}, we
arrive at the following inference algorithm for answering a query $Q$,
using a weight vector $\weightvec$.  Below, we let $N_{v_0}(u)$ denote
the \trm{neighbors} of $u$---i.e., the set of nodes $v$ where
$\Pr(v|u)>0$ (including the restart node $v=v_0$).  We also let
\textbf{W} be a matrix such that
$\textbf{W}[u,v]=\Pr_\weightvec(v|u)$, and in our discussion, we use
$\textbf{ppr}(v_0)$ to denote the personalized PageRank vector for
$v_0$.
\begin{enumerate}
\item Let $v_0=(Q,Q)$ be the start node of the search graph.
  Let $G$ be a graph containing just $v_0$. Let $\textbf{v}^0 = \{ v_0 \}$.
\item For $t=1,\ldots,T$ (i.e., until convergence):
  \begin{list}{\quad}{}
  \item For each $u$ with non-zero weight in $\textbf{v}^{t-1}$, 
   and each $v\in{}N_{u+0}(u)$, add $(u,v,\phi_{\edgeuv})$ to $G$ with weight $\Pr_\weightvec(v|u)$, and set 
   $\textbf{v}^{t} = \textbf{W}\cdot\textbf{v}^{t-1}$
  \end{list}
\item At this point $\textbf{v}^T \approx \textbf{ppr}(v_0)$.  Let $S$
  be the set of nodes $(Q\theta,\Box)$ that have empty subgoal lists
  and non-zero weight in $\textbf{v}^T$, and let $Z=\sum_{u\in{}S}
  \textbf{v}^T[u]$.  The final probability for the literal $L=Q\theta$
  is found by extracting these solution nodes $S$, and renormalizing:
  \[ 
     \Pr_\weightvec(L) \equiv \frac{1}{Z} \textbf{v}^T[(L,\Box)]
  \]
\end{enumerate}
For example, given the query $Q=\textit{about(a,Z)}$ and the program
of Table~\ref{tab:proppr}, this procedure would give assign a non-zero
probability to the literals $\textit{about(a,sport)}$ and
$\textit{about(a,fashion)}$, concurrently building the graph of
Figure~\ref{fig:proof}.

Thus far, we have introduced a language quite similar to SLPs.  The
power-iteration PPR computation outlined above corresponds to a
depth-bounded breadth-first search procedure, and the main extension
of ProPPR, relative to SLPs, is the ability to label a clause
application with a feature vector, instead of the clause's identifier.
Below, however, we will discuss a much faster approximate grounding
strategy, which leads to a novel proof strategy, and a parallelizable
weight-learning method.

\subsection{Locally Grounding a Query}

\begin{table*}
\caption{The PageRank-Nibble-Prove algorithm for inference in ProPPR.
  $\alpha'$ is a lower-bound on $\Pr(v_0|u)$ for any node $u$ to be
  added to the graph $\hat{G}$, and $\epsilon$ is the desired degree
  of approximation.}
\label{tab:pr-nibble}
\begin{center}
\begin{minipage}[t]{\textwidth}
\begin{tabbing}1234\=1234\=1234\=1234\=1234\=1234\=1234\=1234\=1234\=\kill
define PageRank-Nibble-Prove($Q$):\\
\>let $\textbf{v} = $PageRank-Nibble$((Q,Q),\alpha',\epsilon)$\\
\>let $S=\{ u:\textbf{p}[u]>u \mbox{~and~} u=(Q\theta,\Box) \}$\\
\>let $Z=\sum_{u\in S} \textbf{p}[u]$\\
\>define $\Pr_\weightvec(L) \equiv \frac{1}{Z} \textbf{v}[(L,\Box)]$\\
end\\
~\\
define PageRank-Nibble($v_0,\alpha',\epsilon$):\\
\>let $\textbf{p} = \textbf{r} = \textbf{0}$, let $\textbf{r}[v_0] = 1$, and let $\hat{G}=\emptyset$\\
\>while $\exists u: \textbf{r}(u)/|N(u)| > \epsilon$ do: \\
\> \> push($u$)\\
\> return $\textbf{p}$\\
end\\
\end{tabbing}\end{minipage}~~~~~~~\begin{minipage}[t]{\textwidth}
\begin{tabbing}1234\=1234\=1234\=1234\=1234\=1234\=1234\=1234\=1234\=\kill
define push($u, \alpha'$):\\
\> \textit{comment: this modifies \textbf{p}, \textbf{r}, and $\hat{G}$}\\
\> $\textbf{p}[u] = \textbf{p}[u] + \alpha'\cdot\textbf{r}[u]$\\
\> $\textbf{r}[u] = \textbf{r}[u] \cdot (1 - \alpha')$\\
\> for $v\in{}N(u)$:\\
\> \> add the edge $(u,v,\phi_{\edgeuv})$ to $\hat{G}$\\
\> \> if $v=v_0$ then \\
\> \> \> $\textbf{r}[v] = \textbf{r}[v] + \Pr(v|u)\textbf{r}[u]$\\
\> \> else \\
\> \> \> $\textbf{r}[v] = \textbf{r}[v] + {(\Pr(v|u)-\alpha')}\textbf{r}[u]$\\
\> endfor\\
end
\end{tabbing}\end{minipage}
\end{center}
\end{table*}

Note that this procedure both performs inference (by computing a
distribution over literals $Q\theta$) and ``grounds'' the query, by
constructing a graph $G$.  ProPPR inference for this query can be
re-done efficiently, by running an ordinary PPR process on $G$. This
is useful for faster weight learning. Unfortunately, the grounding $G$
can be very large: it need not include the entire database, but if $T$
is the number of iterations until convergence for the sample program
of Table~\ref{tab:proppr} on the query $Q=about(d,Y)$, $G$ will
include a node for every page within $T$ hyperlinks of $d$.

To construct a more compact local grounding graph $G$, we adapt an
approximate personalized PageRank method called PageRank-Nibble
\citep{DBLP:conf/focs/AndersenCL06,DBLP:journals/im/AndersenCL08}.  This method has been used for
the problem of \trm{local partitioning}: in local partitioning, the
goal is to find a small, low-conductance
%\footnote{For small subgraphs
%  $G_S$, \trm{conductance} of $G_S$ is the ratio of the weight of all
%  edges exiting $G_S$ to the weight of all edges incident on a node in
%  $G_S$.}  
component $\hat{G}$ of a large graph $G$ that contains a given node
$v$.

The PageRank-Nibble-Prove algorithm is shown in
Table~\ref{tab:pr-nibble}.  It maintains two vectors: $\textbf{p}$, an
approximation to the personalized PageRank vector associated with node
$v_0$, and $\textbf{r}$, a vector of ``residual errors'' in
$\textbf{p}$.  Initially, $\textbf{p}=\emptyset$ and
$\textbf{r}=\{v_0\}$.  The algorithm repeatedly picks a node $u$ with
a large residual error $\textbf{r}[u]$, and reduces this error by
distributing a fraction $\alpha'$ of it to $\textbf{p}[u]$, and the
remaining fraction back to $\textbf{r}[u]$ and
$\textbf{r}[v_1],\ldots,\textbf{r}[v_n]$, where the $v_i$'s are the
neighbors of $u$.  The order in which nodes $u$ are picked does not
matter for the analysis (in our implementation, we follow Prolog's
usual depth-first search as much as possible.)  Relative to
PageRank-Nibble, the main differences are the the use of a lower-bound
on $\alpha$ rather than a fixed restart weight and the construction of
the graph $\hat{G}$.

Although the result stated in Andersen et al holds only for directed
graphs, it can be shown, following their proof technique, that after
each push, $\textbf{p}+\textbf{r}=\textbf{ppr}(v_0)$.  It is also
clear than when PageRank-Nibble terminates, then for any $u$, the
error $\textbf{ppr}(v_0)[u] - \textbf{p}[u]$ is bounded by
$\epsilon{}|N(u)|$: hence, in any graph where $N(u)$ is bounded, a
good approximation can be obtained.  Additionally, we have the
following efficiency bound:

\begin{theorem}[Andersen,Chung,Lang]
Let $u_i$ be the $i$-th node pushed by PageRank-Nibble-Prove.  Then,\\ 
$\sum_{i} |N(u_i)| < \frac{1}{\alpha'\epsilon}$.
\end{theorem}

This can be proved by noting that initially $||\textbf{r}||_1=1$, and
also that $||\textbf{r}||_1$ decreases by at least
$\alpha'\epsilon|N(u_i)|$ on the $i$-th push.  As a direct
consequence we have the following:

\begin{corollary}
The number of edges in the graph $\hat{G}$ produced by PageRank-Nibble-Prove
is no more than $\frac{1}{\alpha'\epsilon}$.
\end{corollary}

Importantly, the bound holds \emph{independent of the size of the full
  database of facts}.  The bound also holds regardless of the size or
loopiness of the full proof graph, so this inference procedure will
work for recursive logic programs.\footnote{For directed graphs, it
  can also be shown
  \citep{DBLP:conf/focs/AndersenCL06,DBLP:journals/im/AndersenCL08}
  that the subgraph $\hat{G}$ is in some sense a ``useful'' subset of
  the full proof space: for an appropriate setting of $\epsilon$, if
  there is a low-conductance subgraph $G_*$ of the full graph that
  contains $v_0$, then $G_*$ will be contained in $\hat{G}$: thus if
  there is a subgraph $G_*$ containing $v_0$ that approximates the
  full graph well, PageRank-Nibble will find (a supergraph of) $G_*$.}

We should emphasize that this approximation result holds for the
individual nodes in the proof tree, not the answers $Q\theta$ to a
query $Q$.  Following SLPs, the probability of an answer $Q\theta$ is
the sum of the weights of all solution nodes that are associated with
$\theta$, so if an answer is associated with $n$ solutions, the error
for its probability estimate with PageRank-Nibble-Prove may be as
large as $n\epsilon$.

To summarize, we have outlined an efficient approximate proof
procedure, which is closely related to personalized PageRank.  As a
side-effect of inference for a query $Q$, this procedure will create a
ground graph $\hat{G}_Q$ on which personalized PageRank can be run directly,
without any (relatively expensive) manipulation of first-order
theorem-proving constructs such as clauses or logical variables.  As
we will see, this ``locally grounded'' graph will be very useful in
learning weights $\weightvec$ to assign to the features of a ProPPR
program.

\subsection{Learning for ProPPR}

As noted above, inference for a query $Q$ in ProPPR is based on a
personalized PageRank process over the graph associated with the SLD
proof of a query goal $G$.  More specifically, the edges $\edgeuv$ of
the graph $G$ are annotated with feature vectors $\phi_{\edgeuv}$, and
from these feature vectors, weights are computed using a parameter
vector $\weightvec$, and finally normalized to form a probability
distribution over the neighbors of $u$.  The ``grounded'' version of
inference is thus a personalized PageRank process over a graph with
feature-vector annotated edges.

In prior work, Backstrom and Leskovec \citep{backstrom2011supervised} outlined
a family of supervised learning procedures for this sort of annotated
graph.  In the simpler case of their learning procedure, an example is
a triple $(v_0,u,y)$ where $v_0$ is a query node, $u$ is a node in in
the personalized PageRank vector $\textbf{p}_{v_0}$ for $v_0$, $y$ is
a target value, and a loss $\ell(v_0,u,y)$ is incurred if
$\textbf{p}_{v_0}[u] \not= y$.  In the more complex case of ``learning
to rank'', an example is a triple $(v_0,u_+,u_-)$ where $v_0$ is a
query node, $u_+$ and $u_-$ are nodes in in the personalized PageRank
vector $\textbf{p}_{v_0}$ for $v_0$, and a loss is incurred unless
$\textbf{p}_{v_0}[u_+] \geq \textbf{p}_{v_0}[u_-]$.  The core of
Backstrom and Leskovic's result is a method for computing the gradient
of the loss on an example, given a differentiable feature-weighting
function $f(\weightvec,\phi)$ and a differentiable loss function
$\ell$.  The gradient computation is broadly similar to the
power-iteration method for computation of the personalized PageRank
vector for $v_0$. Given the gradient, a number of optimization methods
can be used to compute a local optimum.
 
\begin{sloppypar}
Instead of directly using the above learning approach for ProPPR, 
we decompose the pairwise ranking loss into a standard positive-negative log loss function.  The
training data $D$ is a set of triples $\{
(Q^1,P^1,N^1),\ldots,(Q^m,P^m,N^m)\}$ where each $Q^k$ is a query,
$P^k=\langle Q\theta_+^1,\ldots,Q\theta_+^I\rangle$ is a list of
correct answers, and $N^k$ is a list $\langle
Q\theta_-^1,\ldots,Q\theta_-^J\rangle$ incorrect answers.  
We use a log loss with $L_2$ regularization of the parameter weights.  Hence the final
function to be optimized is
\[
- \Bigg (\sum_{k=1}^I \log \textbf{p}_{v_0}[u_+^k] + \sum_{k=1}^J \log (1 - \textbf{p}_{v_0}[u_-^k]) \Bigg) + \mu||\textbf{w}||^2_2
\]
To optimize this loss, we use stochastic gradient descent (SGD),
rather than the quasi-Newton method of Backstrom and Leskovic.
Weights are initialized to $1.0+\delta$, where $\delta$ is randomly
drawn from $[0,0.01]$.  We set the
learning rate $\beta$ of SGD to be \( \beta =
\frac{\eta}{\mathrm{epoch}^2} \) where epoch is the current epoch in
SGD, and $\eta$, the initial learning rate, defaults to 1.0.
\end{sloppypar}

%Each such
%triple is then locally grounded using the PageRank-Nibble-Prove method
%and used to produce a set $I*J$ of ``learning-to-order'' triples of
%the form $(v^k_0,u_+^{k,i},u_-^{k,j})$ where $v_0^k$ corresponds to
%$Q^k$, and the $u$'s are the nodes in $\hat{G}_{Q^K}$ that correspond
%to the (in)correct answers for $Q^K$. 
%$h=\textbf{p}_{v_0}[u_+] -
%\textbf{p}_{v_0}[u_-]$, i.e.,
%\[
% \ell(v_0,u_+,u_-) \equiv 
% \left\{
    %\begin{array}{cc} h^2 &  \mbox{if $h<0$} \\
    % 0 &  else \\
    %\end{array}
     %\right.
%\]
%WW: rewrote the quadratic loss to log loss
%WW2: there's no definition of learning rate update, so I added here.

We implemented SGD because it is fast and has been adapted to parallel
learning tasks \citep{zinkevich2010parallelized,niu2011hogwild}.
Local grounding means that learning for ProPPR is quite well-suited to
parallelization.  The step of locally grounding each $Q^i$ is
``embarassingly'' parallel, as every grounding can be done
independently.  To parallelize the weight-learning stage, we use
multiple threads, each of which computes the gradient over a single
grounding $\hat{G}_{Q^k}$, and all of which accesses a single shared
parameter vector $\weightvec$. The shared parameter vector is a
potential bottleneck \citep{zinkevich2009slow}; while it is not a
severe one on moderate-size problems, contention for the parameters
becomes increasingly important on the largest tasks we have
experimented with

\section{Inference in a Noisy KB}
In this section, we first introduce the challenges of inference in a noisy KB, and a recently proposed statistical relational learning solution,
then we show how one can apply our proposed locally grounding theory
to improve this learning scheme.
\subsection{Challenges of Inference in a Noisy KB}
A number of recent efforts in industry \citep{singal2012}
and academia \citep{suchanek2007yago,DBLP:conf/aaai/CarlsonBKSHM10,hoffmann2011knowledge} have focused on automatically
constructing large knowledge bases (KBs).  Because
automatically-constructed KBs are typically imperfect and incomplete,
inference in such KBs is non-trivial. 

We situate our study in the context of the NELL (Never Ending Language Learning) research project, which is an effort to develop a
never-ending learning system that operates 24 hours per day, for
years, to continuously improve its ability to read (extract structured
facts from) the web \citep{DBLP:conf/aaai/CarlsonBKSHM10}  NELL is given as
input an ontology that deﬁnes hundreds of categories (e.g., person,
beverage, athlete, sport) and two-place typed relations among these
categories (e.g., athletePlaysSport(Athlete, Sport)), which it must
learn to extract from the web. NELL is also provided a set of 10 to 20
positive seed examples of each such category and relation, along with
a downloaded collection of 500 million web pages from the ClueWeb2009
corpus (Callan and Hoy, 2009) as unlabeled data, and access to 100,000
queries each day to Google’s search engine.  NELL uses a
multi-strategy semi-supervised multi-view learning method to
iteratively grow the set of extracted ``beliefs''.

This task is challenging for two reasons. First, the extensional
knowledge, inference is based on, is not only incomplete, but also
noisy, since its extracted imperfectly from the web. For example, a
football team might be wrongly recognized as two separate entities,
one with connections to its team members, and the other with a
connection to its home stadium.  Second, the size of inference
problems are much larger than those of traditional logical programming
tasks. Given the very large broad-coverage KBs that modern information
extraction systems produce
\citep{suchanek2007yago,DBLP:conf/aaai/CarlsonBKSHM10}, even a
grounding of size that is only linear in the number of facts in the
database, $|DB|$, would impractically large for inference on
real-world problems.

Past work on first-order reasoning has sought to address the first
problem by learning ``soft'' inference procedures, which are more
reliable than ``hard'' inference rules, and address the second problem
by learning restricted inference procedures. In the next sub-section,
we will recap a recent development in solving these problems, and
draws a connection to the ProPPR language.

\subsection{Inference using the \underline{P}ath \underline{R}anking
\underline{A}lgorithm (PRA)}

\citet{DBLP:conf/emnlp/LaoMC11} use the path ranking algorithm (PRA)
to learn an ``inference'' procedure based on a weighted combination of
``paths'' through the KB graph.  PRA is a relational learning system
which generates (and appropriately weights) rules, which accurately
infer new facts from the existing facts in the noisy knowledge base.
As an illustration, PRA's might learn rules such as those in
Table~\ref{tab:rules}, which correspond closely to Horn clauses, as
shown in the Table.  

PRA only considers rules which correspond to ``paths'', or chains of
binary, function-free predicates. Like ProPPR, PRA will weight some
solutions to these paths are weighted more heavily than others:
specifically, weights of the solutions to a PRA ``path'' are based on
random-walk probabilities in the corresponding graph.  For instance,
the last clause of Table~\ref{tab:rules}, which corresponds to the
``path''
\[ 
T \xrightarrow{\mbox{teamHasAthlete}} A\xrightarrow{\mbox{athletePlaysSport}} S
\]
can be understood
as follows:
\begin{enumerate}
\item Given a team $T$, construct a uniform distribution $\mathcal{A}$ of athletes such
that $A \in {\mathcal A}$ is a athlete playing for team $T$.
\item Given ${\mathcal A}$, construct a distribution of sports ${\mathcal S}$ such that $S \in {\mathcal S}$ is played by $A$.  
\end{enumerate}
This final distribution ${\mathcal S}$ is the result: thus the path
gives a weighted distribution over possible sports played by a team.
For a one-clause program, this distribution corresponds precisely to
the distribution produced by ProPPR.

More generally, the output of PRA corresponds roughly to a ProPPR
program in a particular form---namely, the form
\[
\begin{array}{l}
p(S,T) \leftarrow r_{1,1}S,X_1), r_{1,2}(X_1,X_2), \ldots, r_{1,k_1}(X_{k_1-1},T). \\
p(S,T) \leftarrow r_{2,1}(S,X_1), r_{2,2}(X_1,X_2), \ldots, r_{2,k_2}(X_{k_2-1},T). \\
\vdots \\
\end{array}
\]
where $p$ is the binary predicate being learned, and the $r_{i,j}$'s
are other predicates defined in the database.  (In
Table~\ref{tab:rules}, we emphasize that the $r_{i,j}$'s are already
defined by prefixing them with the string ``fact''.)  PRA generates a
very large number of such rules, and then combines them using a sparse
linear weighting scheme, where the (weighted) solutions associated
with a single ``path clause'' are combined with a second set of
weights to produce a final ranking over entity pairs.  More formally,
following the notation of~\citep{DBLP:journals/ml/LaoC10}, define a \trm{relation path}
$P$ as a sequence of relations $r_1, ..., r_{\ell}$. For any relation
path $P = r_1, ..., r_{\ell}$, and seed node $s$, a \trm{path
  constrained random walk} defines a distribution $h$ as $h_{s,P}(e) =
1$ if $e=s$, and $h_{s,p}(e)=0$ otherwise.  If $P$ is not empty, then
$P' = r_1, ..., r_{\ell - 1}$, such that:
\begin{align}
h_{s,P}(e) & = \sum_{e' \in P'} h_{s,P'} (e') \cdot P(e|e';r_{\ell})
\end{align}
where the term $P(e|e';r_{\ell})$ is the probability of reaching node
$e$ from node $e'$ with a one-step random walk with edge type
$r_{\ell}$; that is, it is ${frac}{1}{k}$, where $k = |\{ e' :
r_\ell(e,e') \}|$, i.e., the number of entities $e'$ related to $e$
via the relation $r_\ell$.

Assume we have a set of paths $P_1, ..., P_n$. The PRA algorithm
treats each entity-pair $h_{s,P}(e)$ as a \trm{path feature} for node
$e$, and rank entities using a linear weighting scheme:
\begin{align}
w_1 h_{s,P_1}(e) + w_2 h_{s,P_2}(e) + ... + w_n h_{s,P_n}(e) 
\end{align}
where $w_i$ is the weight for the path $P_i$. PRA then learns the
weights ${\textbf w}$ by performing using elastic net-like regularized
maximum likelihood estimation of the following objective function:
\begin{align}
\sum_i j_i(w) - \mu_1|\!|{\textbf w}|\!|_1 - \mu_2 |\!|{\textbf w}|\!|_2^2
\end{align}
Here $\mu_1$ and $ \mu_2$ are regularization coefficients for elastic
net regularization, and $j_i(w)$ is the per-instance objective
function.  The regularization on $|\!|{\textbf w}|\!|_1$ tends to
drive weights to zero, which allows PRA to produce a sparse classifier
with relatively small number of path clauses.  More details on PRA can
be found elsewhere~\citep{DBLP:journals/ml/LaoC10}.

\begin{center}
\begin{table*}[t] 
\small
\centering % used for centering table 
\hfill{}
\caption{Example PRA rules learned from NELL, written as Prolog clauses.} % title of Table
\label{tab:rules} % is used to refer this table in the text
\begin{tabular}{l c} 
\hline\hline 
\vspace{1ex}
\emph{PRA} Paths for inferring \textbf{athletePlaysSport}:\\
  athletePlaysSport(A,S) :- factAthletePlaysForTeam(A,T),factTeamPlaysSport(T,S).\\
\hline%inserts single line 
\vspace{1ex}
\emph{PRA} Paths for inferring \textbf{teamPlaysSport}:\\
  teamPlaysSport(T,S) :- \\
  \hspace{12 mm}factMemberOfConference(T,C),factConferenceHasMember(C,T'),factTeamPlaysSport(T',S).\\
  teamPlaysSport(T,S) :- \\
  \hspace{12 mm}factTeamHasAthlete(T,A),factAthletePlaysSport(A,S).\\
\hline%inserts single line 
\hline
\end{tabular}
\hfill{}
\end{table*}
\end{center}

\begin{center}
\begin{table*}[t] 
\small
\centering % used for centering table 
\hfill{}
\caption{Example recursive Prolog rules constructed from PRA paths.} % title of Table
\label{tab:recursive-rules} % is used to refer this table in the text
\begin{tabular}{l c} 
\hline\hline 
\vspace{1ex}
Rules for inferring \textbf{athletePlaysSport}:\\
  athletePlaysSport(A,S) :- factAthletePlaysSport(A,S).\\
  athletePlaysSport(A,S) :- athletePlaysForTeam(A,T),teamPlaysSport(T,S).\\
\hline%inserts single line 
\vspace{1ex}
Rules for inferring \textbf{teamPlaysSport}:\\
  teamPlaysSport(T,S) :- factTeamPlaysSport(T,S).\\
  teamPlaysSport(T,S) :- memberOfConference(T,C),conferenceHasMember(C,T'),teamPlaysSport(T',S).\\
  teamPlaysSport(T,S) :- teamHasAthlete(T,A),athletePlaysSport(A,S).\\
\hline%inserts single line 
\hline
\end{tabular}
\hfill{}
\end{table*}
\end{center}

\subsection{From Non-Recursive to Recursive Theories: Joint Inference for Multiple Relations}

One important limitation of PRA is that it learns only programs in the
limited form given above. In particular, PRA can not learn (or even
execute) recursive programs, or programs with predicates of arity more
than two.  PRA also must learn each predicate definition completely
independently.

To see why this is a limitation consider the program in
table~\ref{tab:rules}, which could be learned by PRA by invoking it
twice, once for the predicate \textit{athletePlaysSport} and once for
\textit{teamPlaysSport}.  We call this formulation the
\trm{non-recursive formulation} for a theory.  An alternative would be
to define two mutually recursive predicates, as in
Table~\ref{tab:recursive-rules}.  We call this the \trm{recursive
  formulation}.  Learning weights for theories written using the
recursive formulation is a \emph{joint} learning task, since several
predicates are considered together.  In the next section, we ask the
question: can joint learning, via weight-learning of mutually
recursive programs of this sort, improve performance for a learned
inference scheme for a KB?

\section{Experiments in KB Inference}

To understand the locally groundable first-order logic in depth, we
investigate ProPPR on the difficult problem of drawing reliable
inferences from imperfectly extracted knowledge.  In this experiment,
we create training data by using NELL's KB as of iteration 713, and
test, using as positive examples new facts learned by NELL in later
iterations. Negative examples are created by sampling beliefs from
relations that are mutually exclusive relations with the target
relation. Throughout this section, we set the number of SGD 
optimization epochs to 10.  
Since PRA has already applied the elastic net regularizer when 
learning the weights of different rules, and we are working with
multiple subsets with various sizes of input,
$\mu$ was set to 0 in ProPPR's SGD learning in this section.

For experimentally purposes, we constructed a number of varying-sized
versions of the KB using the following procedure. First, we construct
a ``knowledge graph'', where the nodes are entities and the edges are
the binary predicates from NELL.  Then, we pick a seed entity $s$, and
find the $M$ entities that are ranked highest using a simple untyped
random walk with restart over the full knowledge graph from seed $s$.
Finally, we project the KB to just these $M$ entities: i.e., we select
all entities in this set, and all unary and binary relationships from
the original KB that concern only these $M$ entities.  

This process leads to a coherent, well-connected knowledge base of
bounded size, and by picking different seeds $s$, we can create
multiple different knowledge bases to experiment on.  In the
experiments below, we used the seeds ``Google'', ``The Beatles'', and
``Baseball'' obtaining KBs focused on technology, music, and sports,
respectively.

In this section, we mainly look at three types of rules: 
\begin{itemize}
\item \emph{KB non-recursive}: the simple non-recursive KB rules
that does not contain PRA paths\\ (e.g. teamPlaysSport(T,S) :- factTeamPlaysSport(T,S).);
\item \emph{PRA non-recursive}: the non-recursive PRA rules (e.g. rules in Table~\ref{tab:rules});
\item \emph{PRA recursive}: the recursive formulation of PRA rules (e.g. rules in Table~\ref{tab:recursive-rules}).
\end{itemize}

Since there is currently no structure-learning component for ProPPR,
we construct a program by taking the top-weighted $k$ rules produced
PRA for each relation, for some value of $k$, and then syntactically
transforming them into ProPPR programs, using either the recursive or
non-recursive formulation, as described above.  Again, note that the
recursive formulation allows us to do joint inference on all the
learned PRA rules for all relations at once.

\subsection{Varying The Size of The Graph}

To explore the scalability of the system on large tasks, we evaluated
the performance of ProPPR on NELL KB subsets that have $M=100,000$ and
$M=1,000,000$ entities.  On the 100K subsets, we have 234, 180, and
237 non-recursive KB rules, and 534, 430, and 540 non-recursive/recursive PRA rules in the
Google, Beatles, and Baseball KBs, respectively.  On the 1M subsets,
we have 257, 253, and 255 non-recursive KB rules, and 569, 563, and 567
non-recursive/recursive PRA rules for the three KBs. We set $\epsilon=0.01$ and
$\alpha=0.1$.  Note that we use the top $k=1$ paths to construct
ProPPR programs in the experiments in this subsection.

First we examine the AUC of non-recursive KB rules, non-recursive PRA and recursive PRA ProPPR
theories, after weight-learning, on the 100K and 1M subsets.  From the
table~\ref{tab:AUClarge}, we see that the recursive formulations
performs better in all subsets.  Performance on the 1M KBs are similar, because
the KBs largely overlap (this version of the NELL KB has a little more
than one million entities involved in binary relations.)
When examining the learned weights of the recursive program,
we notice that the top-ranked rules are the recursive PRA rules,
as what we expected.

\begin{table}
\caption{Comparing the learning algorithm's AUC among non-recursive KB, non-recursive PRA, and recursive formulation of ProPPR on NELL 100K and 1M datasets.} \label{tab:AUClarge}
\begin{center}
\begin{tabular}{lrrr}
Methods & Google & Beatles & Baseball \\
\hline
ProPPR 100K KB non-recursive          &0.699     &0.679      & 0.694\\
ProPPR 100K PRA non-recursive       & 0.942    & 0.881    & 0.943\\
ProPPR 100K PRA recursive              & 0.950     & 0.884      &0.952\\
\hline
ProPPR 1M KB non-recursive    &0.701     &0.701     &0.700\\
ProPPR 1M PRA non-recursive &0.945  & 0.944 & 0.945\\
ProPPR 1M PRA recursive        & 0.955     & 0.955     &0.955\\
\hline
\vspace{-5ex}
\end{tabular}
\end{center}
\end{table}

In the second experiment, we consider the training time for ProPPR,
and in particular, how multithreaded SGD training affects the training
time?  Table~\ref{tab:SGDlarge} shows the runtime for the
multithreaded SGD on the NELL 100K and 1M datasets.  Learning takes
less than two minute for all the data sets, even on a single
processor, and multithreading reduces this to less than 20 seconds.
Hence, although we have not observed perfect speedup (probably due to
parameter-vector contention) it is clear that SGD is fast, and that
parallel SGD can significantly reduce the training time for ProPPR.

\iffalse
\begin{figure}[t]
\centerline{\includegraphics[scale=0.28]{./google100kspeedup.png}~~\includegraphics[scale=0.28]{./beatles100kspeedup.png}~~\includegraphics[scale=0.28]{./baseball100kspeedup.png}}
\centerline{\includegraphics[scale=0.28]{./google1Mspeedup.png}~~\includegraphics[scale=0.28]{./beatles1Mspeedup.png}~~\includegraphics[scale=0.28]{./baseball1Mspeedup.png}}
\caption{Performance of the parallel SGD method on NELL 100K and 1M datasets.  The $x$ axis is the
  number of threads on a multicore machine, and the $y$ axis is the
  speedup factor over a single-threaded implementation.
  Top row: 100K datasets. Bottom row: 1M datasets.
  Left column: Google, Middle column: Beatles, Right column: Baseball. }
\label{fig:speeduplarge}
\end{figure}
\fi

\begin{table}
\caption{Runtime (seconds) for parallel SGD of recursive formulation of ProPPR on NELL 100K and 1M datasets.} \label{tab:SGDlarge}
\begin{center}
\begin{tabular}{lrrr}
100K & & &\\
\hline
\#Threads & Google & Beatles & Baseball \\
\hline
1 & 54.9 &   20.0&     51.4\\
2 & 29.4 &   12.1&     26.6\\
4 & 19.1 &   7.4  &    16.8\\
8 & 12.1 &   6.3  &    13.0\\
16  & 9.6   &  5.3  &  9.2\\
\hline
\\
1M & & &\\
\hline
\#Threads & Google & Beatles & Baseball \\
\hline
1        & 116.4 &  87.3  & 111.7\\
2   &  52.6&    54.0  & 59.4\\
4 &  31.0&    33.0  & 31.3\\
8 &  19.0&    21.4   &19.1 \\
16        &15.0  &  17.8   &15.7\\
\hline
\vspace{-5ex}
\end{tabular}
\end{center}
\end{table}

\subsection{Comparing ProPPR and MLNs}
\begin{figure}[t]
\centerline{\includegraphics[scale=0.28]{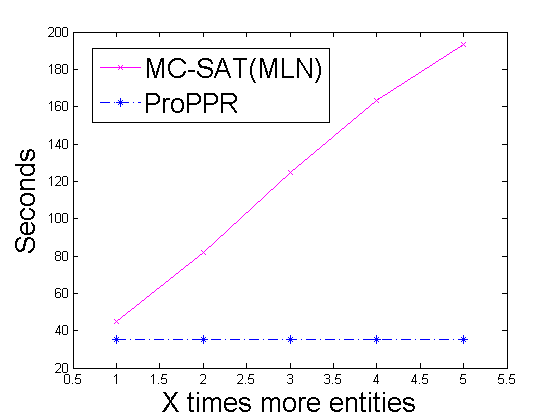}~~\includegraphics[scale=0.28]{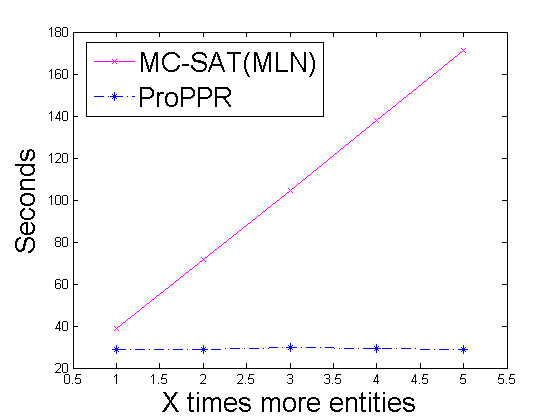}~~\includegraphics[scale=0.28]{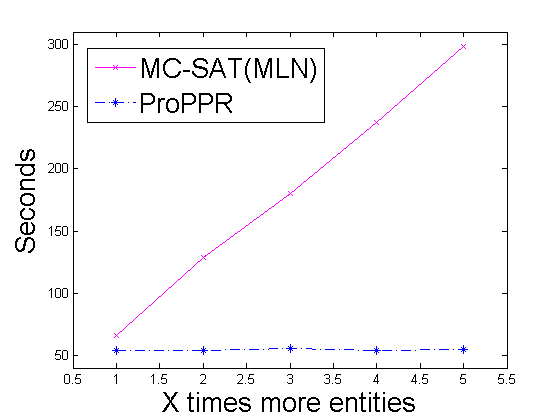}}
\caption{Run-time for non-recursive KB inference on NELL 10K subsets the using ProPPR (with a single
  thread) as a function of increasing the total entities by X times in the database.
 Total test queries are fixed in each subdomain.
  Left, the Google 10K dataset; middle, the Beatles 10K dataset; right, the Baseball 10K
  dataset.}
\label{fig:infer10k}
\end{figure}

Next we quantitatively compare ProPPR's inference time, learning time,
and performance with MLN, using the Alchemy
toolkit.\footnote{http://alchemy.cs.washington.edu/.} We use a KB with
$M=1000$ entities\footnote{We were unable to train MLNs with more than 1,000
  entities.}, and test with a KB with $M=10,000$.  The number
of non-recursive KB rules is 95, 10, and 56 respectively, and the
corresponding number of non-recursive/recursive PRA rules are 230, 29, and 148. The
number of training queries are 466, 520, and 130, and the number of
testing queries are 3143, 2552, and 4906. We set $\epsilon=0.01$ and
$\alpha=0.1$.  Again, we only take the top-1 PRA paths to construct
ProPPR programs in this subsection.

In the first experiment, we investigate whether inference in ProPPR is
sensitive to the size of graph. Using MLNs and ProPPR non-recursive KB
programs trained on the 1K training subsets, we measure evaluate the
inference time on the 10K testing subsets by varying the amount of
entities in the database used at evaluation time. (Specifically, we
use a fixed number of test queries, and increase the total number of
entities in the KB by a factor of $X$, for various values of $X$.)  In
Figure.~\ref{fig:infer10k}, we see that ProPPR's runtime is
independent of the size of the KB. In contrast, when comparing to
MC-SAT, the default (and most efficient) inference method in MLN, we
observe that inference time slows significantly when the database size
grows.

In the second experiment, we compare ProPPR's SGD training method with
MLNs most efficient discriminative learning methods (voted perceptron
and conjugate gradient)~\citep{lowd2007efficient}.  To do this, we
fixed the number of iterations of discriminative training in MLN to
10, and also fixed the number of SGD passes in ProPPR to 10.  In
Table~\ref{tab:train1K}, we show the runtime of various approaches on
the three NELL subdomains. When running on the non-recursive KB theory,
ProPPR has averages 1-2 seconds runtime across all domains, whereas
training MLNs takes hours. When training on the non-recursive/recursive PRA theory,
ProPPR is still efficient.\footnote{We were unable to train MLNs with
  non-recursive or recursive PRA rules.}

\begin{table}
\caption{Comparing the learning algorithm's runtime between ProPPR and MLNs on the NELL 1K subsets .} \label{tab:train1K}
\begin{center}
\begin{tabular}{lrrr}
Method & Google & Beatles & Baseball \\
\hline
ProPPR SGD KB non-recursive    & 2.6       &   2.3       &   1.5\\ 
MLN Conjugate Gradident    & 8604.3 &  1177.4  &   5172.9\\ 
MLN Voted Perceptron       & 8581.4   &  967.3     &  4194.5\\  
\hline
ProPPR SGD PRA non-recursive       &2.6         &3.4          &1.7\\ 
ProPPR SGD PRA recursive       & 4.7         &  3.5         &  2.1\\
\hline
\vspace{-5ex}
\end{tabular}
\end{center}
\end{table}

We now examine the accuracy of ProPPR, in particular, the recursive
formulation, and compare with MLN's popular discriminative learning
methods: voted perceptron and conjugate gradient.  Here, we use AUC of
the ROC curve as the measure. In Table~\ref{tab:AUC1K}, we see that
MLNs outperform ProPPR's using the non-recursive formulation.
However, ProPPR's recursive formulation outperforms all other methods,
and shows the benefits of joint inference with recursive theories.

We should emphasize that the use of AUC means that we are evaluating
only the \emph{ranking} of the possible answers to a query; in other
words, we are not measuring the quality of the actual probability
scores produced by ProPPR, only the relative scores for a particular
query.  ProPPR's random-walk scores tend to be very small for all
potential answers, and are not well-suited to estimating probabilities
in its current implementation.

\begin{table}
\caption{Comparing the learning algorithm's AUC between recursive formulation of ProPPR and MLNs.} \label{tab:AUC1K}
\begin{center}
\begin{tabular}{lrrr}
Methods & Google & Beatles & Baseball \\
\hline
ProPPR SGD KB non-recursive    &0.568     & 0.510      &0.652\\
MLN Conjugate Gradident    & 0.716     & 0.544     & 0.645\\
MLN Voted Perceptron       & 0.826     & 0.573      &0.672\\
ProPPR SGD PRA non-recursive &0.894     & 0.922      &0.930\\ 
ProPPR SGD PRA recursive       & 0.899     & 0.899      &0.935\\
\hline
\vspace{-5ex}
\end{tabular}
\end{center}
\end{table}

\subsection{Varying The Size of The Theory}

\begin{table}
\caption{AUCs for using top-k PRA paths for recursive formulation of ProPPR on NELL 100K and 1M datasets.} \label{tab:AUCtopk}
\begin{center}
\begin{tabular}{lrrr}
Methods & Google & Beatles & Baseball \\
\hline
ProPPR 100K top-1 recursive    & 0.950 &  0.884 &  0.952\\
ProPPR 100K top-2 recursive    & 0.954 &  0.916 &  0.950\\
ProPPR 100K top-3 recursive    & 0.959 &  0.953 & 0.952\\
\hline
ProPPR 1M top-1 recursive    & 0.955 &  0.955 &  0.955\\
ProPPR 1M top-2 recursive    & 0.961 &  0.960 &  0.960\\
ProPPR 1M top-3 recursive    & 0.964 &  0.964 & 0.964\\
\hline
\vspace{-5ex}
\end{tabular}
\end{center}
\end{table}

So far, we have observed improved performance using the recursive
theories of ProPPR, constructed from top $k=1$ PRA paths for each
relation. Here we consider further increasing the size of the ProPPR
program by including more PRA rules in the theory.  In particular, we
also extract the top-2 and top-3 PRA paths (limiting ourselves to
rules with positive weights). On the 100K datasets, this increased the
number of clauses in the recursive theories to 759, 624, and 765 in
the Google, Beatles, and Baseball subdomains in the top-2 condition,
and to 972, 806, and 983 in the top-3 condition.  On the 1M datasets,
we have now 801, 794, and 799 clauses in the top-2 case, and 1026,
1018, and 1024 in the top-3 setup.  From Table~\ref{tab:AUCtopk}, we
observe that using more PRA paths improves performance on all three
subdomains.

%\wc{add the details: how the KB subsets are created, size of theories,
%  ....}
%WW: done

\section{Experiments on Other tasks}
\label{subsec:tasks}

As a further test of generality, we now present results using ProPPR
on two other, smaller tasks.  Our first sample task is an entity
resolution task previously studied as a test case for MLNs
\citep{singla2006entity}.  The program we use in the experiments is
shown in Table~\ref{tab:erprog}: it is approximately the same as the
MLN(B+T) approach from Singla and Domingos.\footnote{The principle
  difference is that we do not include tests on the absence of words
  in a field in our clauses, and we drop the non-horn clauses from
  their program.}  To evaluate accuracy, we use the Cora dataset, a
collection of 1295 bibliography citations that refer to 132 distinct
papers.  
%\wc{mu and num epochs not not listed in experiments above} 
%WW: done.
We set the regularization coefficient $\mu$ to $0.001$ and the number of
epochs to 5.

Our second task is a bag-of-words classification task, which was
previously studied as a test case for both ProbLog \citep{gutmann2010parameter}
and MLNs \citep{lowd2007efficient}. In this experiment, we use the 
following ProPPR program:
\begin{center}
\parbox{0cm}{\begin{tabbing}
class(X,Y) :-\, has(X,W), isLabel(Y), related(W,Y).\\
related(W,Y) :-\, true \# w(W,Y).
\end{tabbing}}
\end{center}
which is a bag-of-words classifier that is approximately\footnote{Note
  that we do not use the negation rule and the link rule from Lowd and
  Domingos.}  the same as the ones used in prior work
\citep{gutmann2010parameter,lowd2007efficient}.  The dataset we use is
the WebKb dataset, which includes a set of web pages from four
computer science departments (Cornell, Wisconsin, Washington, and
Texas).  Each web page has one or multiple labels: \emph{course,
  department, faculty, person, research project, staff, and
  student}. The task is to classify the given URL into the above
categories. This dataset has a total of 4165 web pages. Using our
ProPPR program, we learn a separate weight for each word for each
label.

\begin{table*}
\caption{ProPPR program used for entity resolution.} \label{tab:erprog}
\begin{center}
\begin{small}
\begin{tabular}{|ll|}
\hline
samebib(BC1,BC2) :- &\\
\hspace{10 mm}author(BC1,A1),sameauthor(A1,A2),authorinverse(A2,BC2) & \# author.\\
samebib(BC1,BC2) :- &\\
\hspace{10 mm}title(BC1,A1),sametitle(A1,A2),titleinverse(A2,BC2) & \# title.\\
samebib(BC1,BC2) :- &\\
\hspace{10 mm}venue(BC1,A1),samevenue(A1,A2),venueinverse(A2,BC2) & \# venue.\\
samebib(BC1,BC2) :- &\\
\hspace{10 mm}samebib(BC1,BC3),samebib(BC3,BC2) & \# tcbib.\\ %\hline
% & \\
sameauthor(A1,A2) :- &\\
\hspace{10 mm}haswordauthor(A1,W),haswordauthorinverse(W,A2),keyauthorword(W) & \# authorword.\\
sameauthor(A1,A2) :- & \\
\hspace{10 mm}sameauthor(A1,A3),sameauthor(A3,A2) & \# tcauthor.\\ %\hline
%& \\
sametitle(A1,A2) :- &\\
\hspace{10 mm}haswordtitle(A1,W),haswordtitleinverse(W,A2),keytitleword(W) & \# titleword.\\
sametitle(A1,A2) :- &\\
\hspace{10 mm}sametitle(A1,A3),sametitle(A3,A2) & \# tctitle.\\ %\hline
%& \\
samevenue(A1,A2) :- &\\
\hspace{10 mm}haswordvenue(A1,W),haswordvenueinverse(W,A2),keyvenueword(W) & \# venueword.\\
samevenue(A1,A2) :- &\\
\hspace{10 mm}samevenue(A1,A3),samevenue(A3,A2) & \# tcvenue.\\ %\hline
%& \\
keyauthorword(W) :- true & \#  authorWord(W). \\
keytitleword(W) :- true & \#  titleWord(W). \\
keyvenueword(W) :- true & \#  venueWord(W). \\
\hline
\end{tabular}
\end{small}
\end{center}
\end{table*}

% untrained ProPPR on 52 problems
%epsilon      map                 time(s)
%0.001      0.0377358490566       7           
%0.0001     0.297480954626        28
%0.00005      0.39697867474       39
%0.00002      0.532478145586        75
%0.00001      0.539898788814        116
%0.000005    0.542811175823       216
%ppr      0.540505233578        819

\begin{table}
\caption{Performance of the approximate PageRank-Nibble-Prove method on the Cora dataset, compared to the grounding by running personalized PageRank to
  convergence (power iteration).  In all cases $\alpha=0.1$.} \label{tab:dprVsPPR}
\begin{center}
\begin{tabular}{lrr}
$\epsilon$ & MAP & Time(sec) \\
\hline
0.0001      & 0.30        & 28 \\
0.00005     & 0.40        & 39 \\
0.00002     & 0.53        & 75 \\
0.00001     & 0.54        & 116 \\
0.000005    & 0.54        & 216 \\
\hline
power iteration
            & 0.54        & 819 \\
\vspace{-5ex}
\end{tabular}
\end{center}
\end{table}

For these smaller problems, we can also evaluate the cost of the
PageRank-Nibble-Prove inference/grounding technique on Cora.
Table~\ref{tab:dprVsPPR} shows the time required for inference (with
uniform weights) for a set of 52 randomly chosen entity-resolution
tasks from the Cora dataset, using a Python implementation of the
theorem-prover.  We report the time in seconds for all 52 tasks, as
well as the mean average precision (MAP) of the scoring for each
query.  It is clear that PageRank-Nibble-Prove offers a substantial
speedup on these problems with little loss in accuracy: on these
problems, the same level of accuracy is achieved in less than a tenth
of the time.

\begin{figure}[t]
\centerline{\includegraphics[scale=0.45]{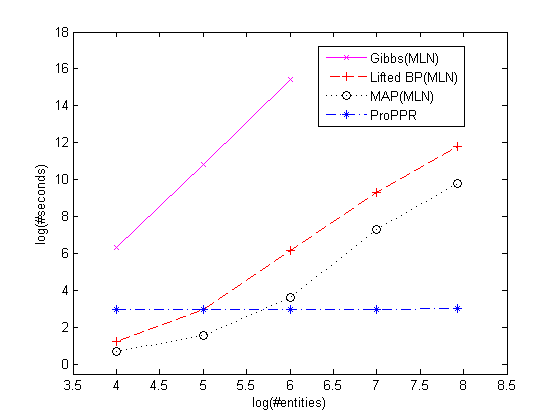}~~\includegraphics[scale=0.28]{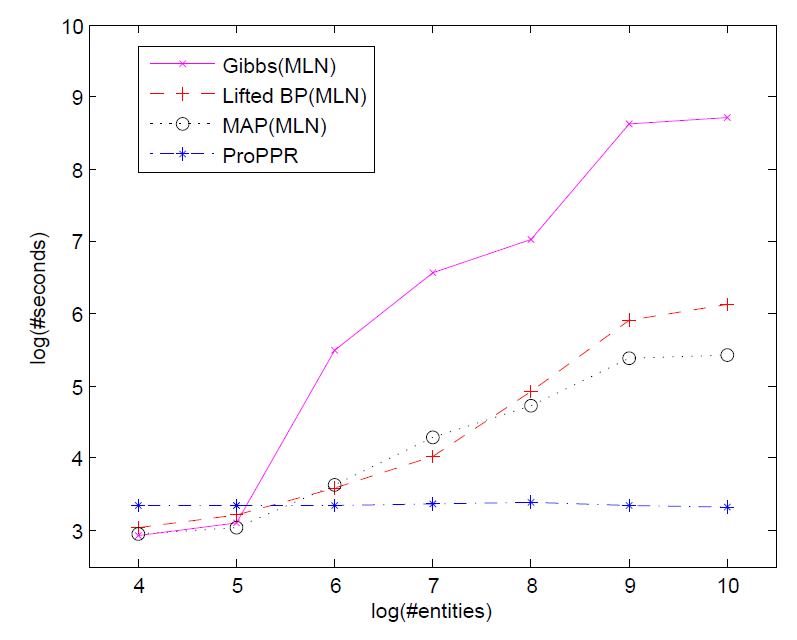}}
\caption{Run-time for inference on the using ProPPR (with a single
  thread) as a function of the number of entities in the database.
  The base of the log is 2. Left, the Cora dataset; right, the WebKB
  dataset.}
\label{fig:time}
\end{figure}

While the speedup in inference time is desirable, the more important
advantages of the local grounding approach are that (1) grounding
time, and hence inference, need not grow with the database size and
(2) learning can be performed in parallel, by using multiple threads
for parallel computations of gradients in SGD.  Figure~\ref{fig:time}
illustrates the first of these points: the scalability of the
PageRank-Nibble-Prove method as database size increases.  For
comparison, we also show the inference time for MLNs with three
inference methods: Gibbs refers to Gibbs sampling,
Lifted BP is the lifted belief propagation method, and MAP is the
maximum a posteriori inference approach.  In each case
the performance task is inference over 16 test queries. 

Note that ProPPR's runtime is constant, independent of the database
size: it takes essentially the same time for $2^8=256$ entities as for
$2^4=16$.  In contrast, lifted belief propagation is up to 1000 times
slower on the larger database.

\begin{figure}[h]
%\wc{can we make the webkb a graph also?}

%\centerline{\fbox{\includegraphics[angle=-90,width=0.56\textwidth]{speedup.png}}} % was: scale=0.3
\includegraphics[angle=-90,width=0.5\textwidth]{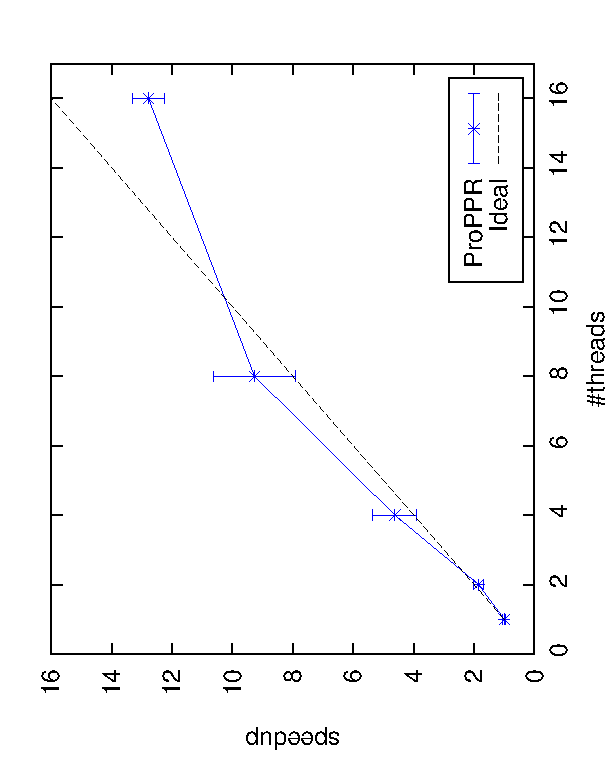}~\begin{tabular}[t]{crrrrr}
~\\~\\~\\~\\
 & Co. & Wi. & Wa. & Te. & Avg. \\ \hline
1 &1190.4 &504.0  &1085.9 &1036.4 &954.2\\
2 &594.9   &274.5 &565.7  &572.5  &501.9\\
4 &380.6  &141.8  &404.2  &396.6  &330.8\\
8 &249.4  &94.5 &170.2  &231.5  &186.4\\ 
16        &137.8  &69.6 &129.6  &141.4  &119.6\\
\hline
\vspace{-5ex}
\end{tabular}
\caption{Performance of the parallel SGD method.  The $x$ axis is the
  number of threads on a multicore machine, and the $y$ axis is the
  speedup factor over a single-threaded implementation.
Left, the Cora dataset; right, the WebKB dataset.}
\label{fig:threads}
\end{figure}

Figure~\ref{fig:threads} explores the speedup in learning (from
grounded examples) due to multi-threading.  The weight-learning is
using a Java implementation of the algorithm which runs over ground
graphs.  For Cora, the speedup is nearly optimal, even with 16 threads
running concurrently. For WebKB, while learning time averages about
950 seconds with a single thread, but this can be reduced to only two
minutes if 16 threads are used. For comparison, Lowd and Domingos
report that around 10,000 seconds were needed to obtain the best
results were obtained for MLNs. 
\begin{table}
\begin{center}
\caption{AUC results on Cora citation-matching.} \label{tab:auc}
\begin{tabular}{lrrrr}
& Cites & Authors & Venues & Titles \\ \hline
MLN(Fig~\ref{tab:proppr})   & 0.513  & 0.532  & 0.602 & 0.544\\
MLN(S\&D)                   & 0.520  & 0.573  & 0.627 & 0.629\\
ProPPR(\textbf{w}=1)        & 0.680  & 0.836  & 0.860 & \textbf{0.908} \\
ProPPR                      & \textbf{0.800}  & \textbf{0.840}  & \textbf{0.869} & 0.900 \\
%WW: is this enough, or should we add other results?
%MLN(B+T)           & 0.950     & 0.994  & 0.745 & $-$ \\
\hline
\end{tabular}
\quad
\caption{AUC results on the WebKb classification task. ProbLog results are from \citep{gutmann2010parameter},
and MLN results are from \citep{lowd2007efficient}.
Co.: Cornell. Wi.: Wisconsin. Wa.: Washington. Te.: Texas.} \label{tab:aucwebkb}
~\\
\begin{tabular}{lrrrrr}
& Co. & Wi. & Wa. & Te. & Avg. \\ \hline
ProbLog   & --  & --  & -- & -- &  0.606\\
MLN (VP)~ & --  & --  & -- & -- & 0.605\\
MLN (CD) & --  & --  & -- & -- & 0.604\\
MLN (CG) & --  & --  & -- & -- & 0.730\\
ProPPR(\textbf{w}=1) & 0.501 & 0.495 & 0.501 & 0.505 & 0.500\\
ProPPR & \textbf{0.785}  & \textbf{0.779}  & \textbf{0.795} & \textbf{0.828}  & \textbf{0.797}\\
\hline
\end{tabular}
\end{center}
\end{table}

We finally consider the effectiveness of weight learning.  For Cora,
we train on the first four sections of the Cora dataset, and report
results on the fifth.  Following Singla and Domingos
\citep{singla2006entity} we report performance as area under the ROC
curve (AUC).  Table~\ref{tab:auc} shows AUC on the test set used by
Singla and Domingos for several methods.  The line for
MLN(Fig~\ref{tab:proppr}) shows results obtained by an MLN version of
the program of Figure~\ref{tab:proppr}.  The line MLN(S\&D) shows
analogous results for the best-performing MLN from
\citep{singla2006entity}.  Compared to these methods, ProPPR does quite
well even before training (with unit feature weights, \textbf{w}=1);
the improvement here is likely due to the ProPPR's bias towards short
proofs, and the tendency of the PPR method to put more weight on
shared words that are rare (and hence have lower fanout in the graph
walk.)  Training ProPPR improves performance on three of the four
tasks, and gives the most improvement on citation-matching, the most
complex task.

The results in Table~\ref{tab:auc} all use the same data and
evaluation procedure, and the MLNs were trained with the
state-of-the-art Alchemy system using the recommended commands for
this data (which is distributed with
Alchemy\footnote{http://alchemy.cs.washington.edu}).  However, we
should note that the MLN results reproduced here are not identical to
previous-reported ones \citep{singla2006entity}. Singla and Domingos
used a number of complex heuristics that are difficult to
reproduce---e.g., one of these was combining MLNs with a heuristic,
TFIDF-based matching procedure based on canopies
\citep{mccallum00efficient}.  While the trained ProPPR model
outperformed the reproduced MLN model in all prediction tasks, it
outperforms the reported results from Singla and Domingos only on
\textit{venue}, and does less well than the reported results on
\textit{citation} and \textit{author}\footnote{Performance on
  \textit{title} matching is not reported by Singla and Domingos.}.

On the Webkb dataset, we use the usual cross-validation method
\citep{lowd2007efficient,gutmann2010parameter}: in each fold, for the
four universities, we train on the three, and report result on the
fourth.  In Table~\ref{tab:aucwebkb}, we show the detailed AUC results
of each fold, as well as the averaged results.  If we do not perform
weight learning, the averaged result is equivalent to a random
baseline.  As reported by Gutmann et al.  the ProbLog approach obtains
an AUC of 0.606 on the dataset \citep{gutmann2010parameter}, and as
reported by Lowd and Domingos, the results for voted perceptron
algorithm (MLN VP, AUC $\approx 0.605$) and the contrastive divergence
algorithm (MLN CD, AUC $\approx 0.604$) are in same range as ProbLog
\citep{lowd2007efficient}. ProPPR obtains an AUC of 0.797, which
outperforms the prior results reported by ProbLog and MLN.

\section{Related work}

Although we have chosen here to compare mainly to MLNs
\citep{RichardsonMLJ2006,singla2006entity}, ProPPR represents a rather
different philosophy toward language design: rather than beginning
with a highly-expressive but intractable logical core, we begin with a
limited logical inference scheme and add to it a minimal set of
extensions that allow probabilistic reasoning, while maintaining
stable, efficient inference and learning.  While ProPPR is less
expressive than MLNs (for instance, it is limited to definite clause
theories) it is also much more efficient.  This philosophy is similar
to that illustrated by probabilistic similarity logic (PSL)
\citep{brocheler2012probabilistic}; however, unlike ProPPR, PSL does
not include a ``local'' grounding procedure, which leads to small
inference problems, even for large databases. Our work also aligns 
with the lifted personalized PageRank \citep{ahmadi2011multi} algorithm, 
which can be easily incorporated as an alternative inference algorithm in our language.

Technically, ProPPR is most similar to stochastic logic programs
(SLPs) \citep{DBLP:journals/ml/Cussens01}.  The key innovation is the
integration of a restart into the random-walk process, which, as we
have seen, leads to very different computational properties.

ProbLog~\citep{de2007problog}, like ProPPR, also supports approximate
inference, in a number of different variants. An extension to ProbLog
also exists which uses decision theoretic analysis to determine when
approximations are acceptable \citep{van2010dtproblog}.  Although this
paper does present a very limited comparison with ProbLog on the WebKB
problem (in Table~\ref{tab:aucwebkb}o) a further comparison of speed
and utility of these different approaches to approximate inference is
an important topic for future work.

There has also been some prior work on reducing the cost of grounding
probabilistic logics: notably, Shavlik et al
\citep{shavlik2009speeding} describe a preprocessing algorithm called
FROG that uses various heuristics to greatly reduce grounding size and
inference cost, and Niu et al \citep{niu2011tuffy} describe a more
efficient bottom-up grounding procedure that uses an RDBMS.  Other
methods that reduce grounding cost and memory usage include ``lifted''
inference methods (e.g., \citep{singla2008lifted}) and ``lazy''
inference methods (e.g., \citep{singla2006memory}); in fact, the
LazySAT inference scheme for Markov networks is broadly similar
algorithmically to PageRank-Nibble-Prove, in that it incrementally
extends a network in the course of theorem-proving.  However, there is
no theoretical analysis of the complexity of these methods, and
experiments with FROG and LazySAT suggest that they still lead to a
groundings that grow with DB size, albeit more slowly.

As noted above, ProPPR is also closely related to the PRA, learning
algorithm for link prediction \citep{DBLP:journals/ml/LaoC10}, like
ProPPR, PRA uses random walk processes to define a distribution,
rather than some other forms of logical inference, such as belief
propagation.  In this respect PRA and ProPPR appear to be unique among
probabilistic learning methods; however, this distinction may not be
as great as it first appears, as it is known there are close
connections between personalized PageRank and traditional
probabilistic inference schemes\footnote{For instance, it is known
  that personalized PageRank can be used to approximate belief
  propagation on certain graphs \citep{cohen2010graph}.}.  PRA,
however, is much more limited than ProPPR, again, as noted above.
However, unlike PRA, we do not consider the task of searching for
logic program clauses.

\section{Conclusions}
\begin{sloppypar}
We described a new probabilistic first-order language which is
designed with the goal of highly efficient inference and rapid
learning.  ProPPR takes Prolog's SLD theorem-proving, extends it with
a probabilistic proof procedure, and then limits this procedure
further, by including a ``restart'' step which biases the system to
short proofs.  This means that ProPPR has a simple polynomial-time
proof procedure, based on the well-studied personalized PageRank (PPR)
method.  

Following prior work on approximate PPR algorithms, we designed a
local grounding procedure for ProPPR, based on local partitioning
methods
\citep{DBLP:conf/focs/AndersenCL06,DBLP:journals/im/AndersenCL08},
which leads to an inference scheme that is an order of magnitude
faster that the conventional power-iteration approach to computing
PPR, takes time $O(\frac{1}{\epsilon\alpha'})$, independent of
database size.  This ability to ``locally ground'' a query also makes
it possible to partition the weight learning task into many separate
gradient computations, one for each training example, leading to a
weight-learning method that can be easily parallelized.  In our
current implementation, an additional order-of-magnitude speedup in
learning is made possible by parallelization.  Experimentally, we
showed that ProPPR performs well on an entity resolution task, and a
classification task.  It also performs well on a difficult problem
involving joint inference over an automatically-constructed KB, an
approach that leads to improvements over learning each
predicate separately.  Most importantly, ProPPR scales well, taking
only a few seconds on a conventional desktop machine to learn weights
for a mutually recursive program with hundreds of clauses, which
define scores of interrelated predicates, over a substantial KB
containing one million entities.
\end{sloppypar}

In future work, we plan to explore additional applications of, and
improvements to, ProPPR.  One improvement would be to extend ProPPR to
include ``hard'' logical predicates, an extension whose semantics have
been fully developed for SLPs \citep{DBLP:journals/ml/Cussens01}.
Also, in the current learning process, the grounding for each query
actually depends on the ProPPR model parameters.  We can potentially
get improvement by making the process of grounding more closely
coupled with the process of parameter learning. Finally, we note that
further speedups in multi-threading might be obtained by incorporating
newly developed approaches to loosely synchronizing parameter updates
for parallel machine learning methods \citep{ho2013more}.  

\subsubsection*{Acknowledgements} 
This work
was sponsored in part by DARPA grant FA87501220342 to CMU and a Google
Research Award.

%\subsubsection*{References} 

%\begin{acknowledgements}
%\end{acknowledgements}

\bibliography{./all}  
\bibliographystyle{spbasic} 

%\begin{thebibliography}{}
% etc
%\end{thebibliography}

\end{document}